\pdfoutput=1

\documentclass[11pt]{article}

\usepackage[]{acl}

\usepackage{times}
\usepackage{latexsym}

\usepackage{booktabs}
\usepackage{amsmath}
\usepackage{amsfonts}
\usepackage{multirow}

\usepackage[T1]{fontenc}

\usepackage[utf8]{inputenc}

\usepackage{microtype}

\usepackage{inconsolata}

\usepackage{graphicx}

\title{Reference-free Evaluation Metrics for Text Generation: A Survey}

\author{
  Takumi Ito
  \\
  Tohoku University 
  \\
  Langsmith Inc.
  \\
  \texttt{t-ito@tohoku.ac.jp}
  \And
  Kees van Deemter 
  \\
  Utrecht University
  \\
  \texttt{c.j.vandeemter@uu.nl}
  \And
  Jun Suzuki 
  \\
  Tohoku University 
  \\
  RIKEN
  \\
  \texttt{jun.suzuki@tohoku.ac.jp}
}

\begin{document}
\maketitle
\begin{abstract}
A number of automatic evaluation metrics have been proposed for natural language generation systems.
The most common approach to automatic evaluation is the use of a reference-based metric that compares the model's output with gold-standard references written by humans.
However, it is expensive to create such references, and for some tasks, such as response generation in dialogue, creating references is not a simple matter.
Therefore, various \textbf{reference-free} metrics have been developed in recent years.
In this survey, which intends to cover the full breadth of all NLG tasks, we investigate the most commonly used approaches, their application, and their other uses beyond evaluating models.
The survey concludes by highlighting some promising directions for future research.
\end{abstract}

\section{Introduction}
The performance of natural language generation (NLG) technology~\citep{Reiter-book} has improved dramatically in recent years~\citep{clark-etal-2021-thats}.
As the errors made by NLG systems are becoming subtler, the detection of errors in NLG models is likewise becoming more complex. 
Consequently, recent years have seen a growing focus on the \emph{evaluation} of NLG models~\citep{sai-2022-survey,celikyilmaz-2021-survey,novikova-etal-2017-need,li2024leveraging}.

The evaluation of NLG models can be categorized into human and automatic evaluation.
Although human evaluation is widely regarded as more convincing than computational metrics~\citep{van-der-lee-etal-2019-best,graham_baldwin_moffat_zobel_2017}, human evaluation tends to be costly.
Therefore, automatic evaluation is often used in practice.
Automatic evaluation can be divided into two types: reference-based and reference-free.
Reference-based metrics evaluate the quality of a text by measuring the correspondence between system outputs and human-written texts, termed references, which are considered gold standards for evaluation.
Reference-based metrics require the preparation of references, and humans must create and verify the reference texts. 
In addition, many NLG tasks require multiple reference texts because the same information may be expressed in many different ways, no one of which is necessarily better than the others.
Moreover, the performance of reference-based methods greatly depends on the quality and quantity of the references~\citep{freitag-etal-2020-bleu}.
If there are few references, or their quality is poor, they will not provide proper evaluation.

To address the limitations of reference-based metrics, reference-free metrics, which evaluate NLG systems without references, have been proposed.\footnote{Reference-free evaluation is also called quality estimation (QE)~\citep{callison-burch-etal-2012-findings,scarton-etal-2016-quality}.}
Reference-free evaluation has the potential to greatly increase the scalability of NLG evaluation.
Moreover, the performance of reference-free evaluation, in terms of the strengths of its correlation with human evaluation ratings, has recently seen improvements~\citep{rei-etal-2021-references,islam-magnani-2021-end}.

While many reference-free approaches have been proposed, most NLG evaluation studies have so far been reference-based.
Unsurprisingly therefore, surveys of evaluation metrics for individual NLG tasks~\citep{chauhan2022comprehensive,ERMAKOVA20191794} and comprehensive surveys of NLG evaluations~\citep{celikyilmaz-2021-survey,sai-2022-survey} have paid little attention to reference-free approaches. 
The present survey intends to fill this gap, paying attention to the full breadth of NLG tasks (Table~\ref{tab:nlg-tasks}).
To clarify the value and limitations of reference-free evaluation metrics and to encourage their future study, we offer a survey of the main approaches to reference-free evaluation~\citep[e.g.,][]{chimoto-bassett-2022-comet} and their analysis~\citep[e.g.,][]{mohiuddin-etal-2021-rethinking,durmus-etal-2022-spurious}.

\section{Terminology}
\label{sec:terminology}

\begin{table*}[t]
\centering
\footnotesize{
\renewcommand{\arraystretch}{1}
    \begin{tabular}{lll} 
    \toprule
    Task &  Context & Hypothesis\\
    \cmidrule(r){1-1} \cmidrule(r){2-2} \cmidrule(r){3-3}
    Machine Translation (MT) & Source language text & Translation \\
    Summarization (SUM) & Document(s) & Summary \\
    Question Answering (QA) & Question + Background knowledge (e.g., Knowledge base, Image) & Answer \\
    Question Generation (QG) & Background knowledge (e.g., Knowledge base, Image) & Question \\
    Dialogue (DG) & Conversation history & Response \\
    Story Generation (SG) & Premises or Summary & Story \\
    Image Captioning (IC) & Image & Caption \\
    Data-to-Text (D2T) & Structured data (e.g., Table, Graph) & Text \\
    \bottomrule
    \end{tabular}
    }
    \caption{Context and hypothesis of typical NLG tasks, modified from Table~2 in ~\citet{sai-2022-survey}.}
    \label{tab:nlg-tasks}
\end{table*}

In this paper, we use the term \textbf{reference} ($r$) for a piece of text that can be considered to be correct, for instance because it was written or validated by humans. 
References are used for assessing the quality of the outputs of the NLG model. 
Outputs are thus viewed as \textbf{hypotheses} ($h$). 
All information used for the evaluation metrics other than hypotheses and references is \textbf{context} ($c$).
Often, the context is the NLG model's input, but not always. 
For example, in a dialogue generation task, some evaluation metrics make use of future utterances, which the NLG model does not have access to~\citep{li-etal-2021-conversations,mehri-eskenazi-2020-unsupervised}. 

Reference-free metrics are typically categorized into two types: absolute evaluation and ranking evaluation. 
Absolute evaluation takes a context ($c$) and a hypothesis ($h$), and returns a quality score.
Ranking evaluation takes a context ($c$) and a set of hypotheses, and ranks the hypotheses; in particular, the method of ranking two hypotheses ($h_1$ and $h_2$) is called pairwise evaluation.

Table~\ref{tab:nlg-tasks} presents contexts and hypotheses for each NLG task.
This table is reconstructed from Table~2 in \citet{sai-2022-survey}, adding the story generation task and revising some of the task settings.
Unless otherwise noted, these formats of contexts and hypotheses for each task are assumed in this paper.

\section{Research questions}
Generally, reference-based approaches use similarity to a reference to estimate the quality of the hypothesis, given the context; we will say that such similarity metrics function as a \emph{proxy} for the quality of the hypothesis (given the context). 
In the absence of references, text quality assessment methods differ along a number of important dimensions, which will be used to organize this survey.

The first question is (i) \textit{How do these approaches evaluate the quality of the hypothesis, and what do they use as a proxy for quality?} (Section~\ref{sec:generic-approach})
While many methods consider both the context and the hypothesis, some focus exclusively on evaluating the hypothesis alone.
Often used supplementary in reference-free evaluation scenarios, these methods offer an approach to assessing textual qualities such as fluency.
This raises the question (ii) \textit{How can hypothesis-only evaluation methods evaluate textual qualities such as fluency?} (Section~\ref{sec:evaluatuion-aspects})
Finally, reference-free metrics can have other uses beyond comparing NLG models' performance.
For example, they can be used in a re-ranking function to select the most optimal hypotheses from an NLG model and in a reward function for reinforcement learning.

They have also been used to filter out low-quality training data of NLG models~\citep{bane-etal-2022-comparison}.
We therefore ask, (iii) \textit{How can reference-free metrics be used, other than for evaluating the quality of the hypothesis in an NLG task?} (Section~\ref{sec:other-usages}). 

\section{What is the proxy for quality?}
\label{sec:generic-approach}

A reference-based method evaluates text quality by comparing the text (i.e., the hypothesis) with the reference.
For reference-free approaches, we discuss how quality is evaluated and what is used as an evaluation proxy.
Some reference-free metrics combine several of the approaches presented below.
Methods that focus solely on the hypothesis, ignoring the context, are discussed in Section~\ref{sec:evaluatuion-aspects}.

\subsection{Learning from human judgments}
\label{subsec:human-judgement}
One approach to constructing reference-free metrics is to use a regression model to predict human judgments given a context and a hypothesis
~\citep{heilman-etal-2014-predicting,xenouleas-etal-2019-sum}.
This idea is straightforward because if an evaluation metric correlates strongly with human judgments, it must be reliable. 
For example, COMET-QE~\citep{rei-etal-2021-references} uses a regression model trained on human scores of MT using features of the context and the hypotheses that are acquired using a cross-lingual pre-trained language model.

\paragraph{Scope and merits.}
This method can be applied to all NLG tasks (Table~\ref{tab:nlg-tasks}), as long as high-quality human judgments are available in large quantities. 
In particular, in the context of MT evaluation, human evaluation data is consistently compiled for the annual shared task of evaluation metrics~\citep{freitag-etal-2021-results,freitag-etal-2022-results,freitag-etal-2023-results}, which has led to the frequent application of this method in MT~\citep{rei-etal-2021-references}.
With a large quantity of high-quality human judgments, it should be possible to construct high-performance evaluation metrics in this way.

\paragraph{Limitation.}
Collecting large amounts of high-quality human judgments is costly.

\subsection{Learning from pseudo-judgments}
\label{subsubsec:psuedo-rating}
Instead of human judgments, pseudo-judgments are likewise used as a proxy. 
The most prevalent method for generating pseudo-ratings involves utilizing scores from reference-based evaluation metrics~\citep{chollampatt-ng-2018-neural,zouhar-etal-2023-poor}.
In this case, generating pseudo-judgments requires reference and hypothesis data, which are often NLG model training data and the corresponding output from an NLG model, respectively.

To generate pseudo-judgments, simple approaches are also sometimes used.
For example, the reference is given a ``good" judgment, while the output of the NLG model or a noised version of the reference (such as with word drops) is assigned a ``bad" judgment~\citep{bao-etal-2022-suenes,guan-huang-2020-union,wu-etal-2020-unsupervised,lee-etal-2021-umic,moosa2024mtranker}.
This approach could be better suited for building pairwise ranking evaluation metrics because a simple ``good or bad'' decision would be sufficient for constructing such metrics, and it makes it easier to generate training data for pairwise comparisons~\citep{moosa2024mtranker}.

\paragraph{Scope and merits.} 
This approach is used for data augmentation to supplement limited human judgments.
Moreover, this approach is used to effectively transforms reference-based evaluation methods into reference-free evaluation metrics, allowing for the benefits of reference-free methods, as shown in Section~\ref{sec:other-usages}.

\paragraph{Limitation.}
It is important to note that discrepancies between pseudo-judgments and actual evaluations are likely.

\subsection{Correspondence between context and hypothesis}\label{subsec:similarity}

``Correspondence" is the extent to which the hypothesis aligns with the context. 
In various NLG tasks, the hypothesis should faithfully reflect the context and accurately convey the necessary details. 
Therefore, the correspondence between the context and hypothesis is often used as a proxy for evaluating these tasks.
We discuss the general scope, merits, and limitations of this approach, followed by a detailed explanation of the method for comparing context and hypotheses.

\paragraph{Scope and merits.}
The correspondence-based approach is often used in machine translation, summarization tasks, image captioning, and data-to-text tasks~\citep{albert-etal-2018-survey-nlg,Reiter-book}.
For example, in machine translation, the output must express the same information as the input; in summarization, the emphasis is on whether the generated summary encapsulates the most important information from the original text; similarly, in image captioning, the emphasis is on accurately describing the main content of the image. 
Information not present in the context should not appear in the hypothesis. 
The correspondence-based approach is used to test these aspects. 
The correspondence-based approach is applied to test the veracity of the hypothesis~\citep{vanDeemter-2024}, as it is commonly employed to detect hallucinations in generated content.
In particular, approaches using Question Answering (QA) and  Natural Language Inference (NLI) are often used for this purpose~\citep{fabbri-etal-2022-qafacteval}, as will be explained below.

\paragraph{Limitations.}
Although these methods can successfully apply to the veracity of generated texts, they are difficult to apply to such quality issues as verbosity, duplication of information, lack of coherence, lack of fluency, and impoliteness.
Other limitations apply to other NLG tasks. 
For example, in story generation, new information, which is not present in the context, often has to be generated, and 
the correspondence-based approach is not suitable for evaluating the quality of such new information.
The correspondence-based approach can only test the presence of information that should (or should not) be included in the hypothesis based on the context.
Therefore, it is often used in combination with other approaches. 
For example, in machine translation, it is used in conjunction with the perplexity of hypotheses as given by language models~\citep{zhao-etal-2020-limitations}.

\subsubsection{Encoding as embedding}
One way to perform correspondence-based evaluation involves encoding both the context and the hypothesis as embeddings, and then assessing the relationship using measures such as cosine similarity or word mover's distance. 

\paragraph{Scope and merits.}
The method is often used in machine translation and image captioning.
XMoverScore~\citep{zhao-etal-2020-limitations}, used in machine translation, and CLIPScore~\citep{hessel-etal-2021-clipscore}, used in image captioning, are examples of evaluation metrics that use this approach. 
XMoverScore utilizes multilingual BERT (MBERT)~\citep{devlin-etal-2019-bert}, which supports multilingual text embeddings, while CLIPScore utilizes CLIP~\citep{pmlr-v139-radford21a}, which can encode both images and text as embeddings.

\paragraph{Limitations.}
An embedding model that consistently places corresponding context and hypothesis close in its vector space is important for this approach. 
However, not all embedding models ensure this.
For example, it has been suggested that the monolingual subspaces of MBERT, which is used as an embedding for machine translation evaluation metrics, do not align well with each other, and require re-mapping~\citep{zhao-etal-2020-limitations,belouadi-2022-uscore}.

\subsubsection{Reverse transformation.}
Another approach is to follow up the transformation performed by NLG by its reverse transformation, then compare the original context with the result of that reverse transformation, then evaluate whether the original context has been correctly restored.

\paragraph{Scope and merits.}
This approach to correspondence-based evaluation is often used in machine translation, where back-translation models are readily available. 
For example, in English-to-German translation (forward translation), a German-to-English translation model is prepared, and then the input English text of the model is compared with the back-translated English text, using reference-based metrics~\citep{moon-etal-2020-revisiting,zhuo-etal-2023-rethinking}.
When using an NLG model for the reverse transformation, it is similar to methods that will be discussed in Section~\ref{subsubsec:sim}.
In other tasks, such as generating text from semantic representations, \citet{manning-schneider-2021-referenceless} a similar approach has been applied by using a parser that converts text to semantic representations to evaluate whether the generation process can be accurately reversed.

\paragraph{Limitations.}
The main limitation of this approach is that it depends on the performance of the reverse transformation~\citep{moon-etal-2020-revisiting,zhuo-etal-2023-rethinking}.

\subsubsection{Using question answering (QA).}
QA is used to evaluate whether a text contains the necessary information.
The idea behind this approach is to check whether the answers obtained by a QA system, when referencing the context and the hypothesis separately for a given question, are the same.
If the answers are same, it is assumed that the information regarding the question is contained in both the context and the hypothesis. 
The critical point in this approach is how to extract suitable questions.

\paragraph{Scope and merits.}
This approach is often applied for the evaluation of summarization.
\citet{eyal-etal-2019-question} proposed a (reference-based) method for identifying important entities from references and generating fill-in-the-blank questions.
Then, \citet{scialom-etal-2019-answers} extended the method to generate fill-in-the-blank questions from the context, and \citet{wang-etal-2020-asking} generated questions from the hypothesis. 
\citet{scialom-etal-2021-questeval} proposed a method for generating questions from both context and hypothesis.
The idea of using QA for testing other NLP systems was used by \citet{lee-etal-2021-qace-asking} in testing a caption generation system and by \citet{rebuffel-etal-2021-data} in testing a table-to-text system.

\paragraph{Limitations.}
The success of this approach depends on two elements: (i) creating good questions~\citep{gabriel-etal-2021-go} and (ii) having a high-performing QA system.

\subsubsection{Using natural language inference (NLI).}
NLI is the task of predicting whether one sentence or text implies or contradicts another (or neither, i.e., ``neutral''). 
NLI-based approach are often used for summarization~\citep{laban-etal-2022-summac} and dialogue evaluation~\citep{dziri-etal-2019-evaluating,pang-etal-2020-towards}. 
For example, in dialogue, it should be noted that each utterance is consistent with previous ones,\footnote{Recall that the \emph{context} of an utterance in dialogue is considered to consist of the utterances preceding it in the dialogue. (See Table~\ref{tab:nlg-tasks})}, and NLI is often used to detect inconsistencies.

\paragraph{Scope and merits.}
NLI-based approaches are often used for summarization~\citep{laban-etal-2022-summac} and dialogue evaluation~\citep{dziri-etal-2019-evaluating,pang-etal-2020-towards}.
For example, in dialogue, it should be noted whether each utterance is consistent with previous ones,\footnote{Recall that the \emph{context} of an utterance in dialogue is considered to consist of the utterances preceding it in the dialogue. (See Table~\ref{tab:nlg-tasks})}, and NLI is often used to detect inconsistencies. 

\paragraph{Limitations.}
A simple way to perform this approach is to use models trained on existing NLI datasets, such as SNLI~\citep{bowman-etal-2015-large} and MultiNLI~\citep{williams-etal-2018-broad}.
However, these existing datasets are known to have biases~\citep{gururangan-etal-2018-annotation}.
In addition, existing NLI datasets may not meet the requirements for NLG evaluation due to factors such as domain mismatch.
Indeed, \citet{falke-etal-2019-ranking} confirmed that models trained on existing NLI datasets cannot be used immediately for summarization evaluation. 
\citet{dziri-etal-2019-evaluating} built NLI models by creating a pseudo-NLI dataset for dialogue with normal conversational data as entailment, unrelated dialogue and dull responses as neutral, and ungrammatical sentences and contradiction data of MuitlNLI as a contradiction.

\subsubsection{Comparing key expressions.}
Lastly, the correspondence-based approach can be performed by
extracting key expressions from the context and hypothesis and comparing them. 
The critical point in this relatively simple approach is how to extract keywords.
For example, KoBE~\citep{gekhman-etal-2020-kobe}, a metric for evaluating machine translation, uses an entity linking system with a multilingual knowledge base to extract entities (keywords) from both the context and hypothesis, and then compares these entities to check correspondences.
In image captioning, \citet{madhyastha-etal-2019-vifidel} proposed a metric that detects an object in an image and compares the labels of detected objects, and the caption using the Word Mover Distance.

\paragraph{Scope and merits.}
This method is good for testing whether or not a certain thing is mentioned.
It is a simple method that is useful for checking for obvious omissions.

\paragraph{Limitations.}
This approach is intrinsically very limited because keyword extraction does not address the full meaning of the hypothesis or the context.
For example, even if the text mentions all the correct objects, adding a negation will change the meaning of the sentence dramatically, potentially turning a good hypothesis into a bad one or conversely.

\subsection{Peer evaluation}
Frequently, one NLG model is used as a proxy for evaluating another.
With some irony, this had been called ``peer" evaluation.

\subsubsection{Probability given other NLG models}
\label{subsubsec:peer-review}
In order to address the correspondence between context and hypothesis,
high-performing NLG models that differ from the model to be evaluated are often used~\citep{fu2023gptscore, thompson-post-2020-automatic}.
For example, Prism-src~\citep{thompson-post-2020-automatic} uses a multilingual machine translation model to calculate the generation probability of the hypothesis for the context and take that probability as the evaluation score.
In addition, \citet{mehri-eskenazi-2020-unsupervised,li-etal-2021-conversations} propose the use of a dialogue model to evaluate other dialogue models. 

\paragraph{Scope and merits.}
This method can be applied to all NLG tasks (Table~\ref{tab:nlg-tasks}) in principle, as long as a high-performance NLG model is available.

\paragraph{Limitations.}
This approach suffers from the issue that a high-performance NLG model is necessary to evaluate the performance of an NLG model~\citep{deutsch-etal-2022-limitations}. 
Meanwhile, \citet{agrawal-etal-2021-assessing} suggest that an evaluation translation model can correctly rank other translation models that outperform it on average.
Further investigation is needed to weigh the strengths and weaknesses of this approach.

\subsubsection{Similarity with pseudo-reference}
\label{subsubsec:sim}
The idea of this approach is to use pseudo-references~\citep{belouadi-2022-uscore,chen-etal-2021-training,gao-etal-2020-supert}.
Here, $\mathrm{Similarity}$ is usually calculated by a reference-based metric.
This approach may be regarded as one of the methods outlined in Section~\ref{subsec:similarity}, as it addresses the challenges of context and hypothesis formulation by employing pseudo-references.

\paragraph{Scope and merits.}
Pseudo-references are generated by NLG models other than the model being evaluated or, in some cases, by heuristic algorithms.
Therefore, the method can be applied to all NLG tasks (Table~\ref{tab:nlg-tasks}), as long as a high-performance NLG model is available.
For example, \citet{belouadi-2022-uscore} propose an evaluation metric for machine translation, part of which creates pseudo-references using another machine translation model and compares the pseudo-references to the output of the test model, using Word Mover Distance.

\paragraph{Limitations.}
Analogous to the reference-based approach, the quality of the pseudo-references significantly influences the evaluation performance. 
A notable challenge for this approach is the necessity of having access to a high-performance NLG model for the effective evaluation of other NLG models.

\subsubsection{LLM-as-a-judge}
\label{subsubsec:llm-as-a-judge}

Recent advancements have seen the increasing utilization of LLMs in the evaluation of NLG tasks, a method commonly referred to as ``LLM-as-a-judge''~\citep{zheng2023judging,chiang-lee-2023-large,kocmi-federmann-2023-large}. 
The LLM-as-a-judge presents evaluation results as text output, such as generating evaluation scores like \texttt{``this hypothesis is 3 score''}. 
The expression in the formula is as follows.

\paragraph{Scope and merits.}
The advantage of LLMs lies in its ability to adapt to various tasks and evaluation criteria through prompting. 
Additionally, LLMs can generate explanation such as the reasoning behind scores. 
Furthermore, the emergence of multi-modal LLMs has shown promising results in tasks such as image caption~\citep{lee-etal-2024-fleur}.

\paragraph{Limitations.}
While it is a promising approach, there are several challenges~\citep{zheng2023judging,chen-etal-2024-humans,ohi-etal-2024-likelihood}.
For example, it has been reported that slight modifications in the prompt can alter evaluations, and there are biases like giving higher ratings to longer hypotheses~\citep{zheng2023judging}. 
Additionally, most LLMs used in this approach are proprietary commercial models (e.g., GPT-4), only accessible via an API, and subject to frequent updates, which poses challenges to reproducibility.

\section{How can textual quality be evaluated?}
\label{sec:evaluatuion-aspects}
Some reference-free metrics are designed to evaluate hypotheses only, without paying attention to context. 
Such metrics are used to evaluate textual quality and they are often combined with other reference-free metrics, such as those described in Section~\ref{subsec:similarity}. 
This section focuses on often discussed textual qualities, such as fluency, and on typical approaches to using these qualities in NLG evaluation.

\subsection{Supervised modeling of textual quality on annotated data}
Research on tasks such as essay scoring, which involve qualities such as fluency and coherence, is actively conducted. 
Models trained for these tasks are also used as evaluation metrics in NLG.
Here, we briefly discuss several datasets in which scores  for fluency and coherency have been manually assigned to texts. 
Datasets of this kind can be used for training models of textual quality, offering significant benefits. 
However, the construction of such datasets involves considerable expense.

\paragraph{Fluency.} 
The Corpus of Linguistic Acceptability (CoLA)~\citep{warstadt-etal-2019-cola} is a dataset comprising English sentences, each annotated as either grammatical or ungrammatical. 
CoLA is often used to train classification models for fluency assessment~\citep{zhu-bhat-2020-gruen,krishna-etal-2020-reformulating}.
Italian CoLA~\citep{trotta-etal-2021-monolingual-cross} and Russian CoLA~\citep{mikhailov-etal-2022-rucola} have also been created.
SummEval~\citep{fabbri-etal-2021-summeval} is a benchmark dataset for summarization evaluation, featuring fluency score annotations applied to the outputs of various summarization models.
Additionally, TMU-GFM-Dataset~\citep{yoshimura-etal-2020-reference} is a dataset specifically designed for Grammatical Error Correction (GEC) metrics. 
This dataset includes annotations for both fluency and grammaticality in the outputs of GEC systems.

\paragraph{Coherence.} 
The Grammarly Corpus of Discourse Coherence (GCDC)~\citep{lai-tetreault-2018-discourse} is a dataset featuring English texts that have been manually annotated to reflect three levels of discourse coherence: low, medium, and high. 
Classification model trained on GCDC is used for evaluating coherence~\citep{vasquez-rodriguez-etal-2023-document}.
Following the GCDC, datasets annotated for discourse coherence in Danish~\citep{flansmose-mikkelsen-etal-2022-ddisco} and Chinese~\citep{wu-etal-2023-multi-task} have subsequently been created. 
SummEval also provides coherence score as well as fluency.

\subsection{Language models}\label{subsec:language-model}

LMs are often used to evaluate the fluency, coherence, and readability of hypotheses, without relying on annotated data, without relying on annotated data~\citep{mehri-eskenazi-2020-usr,wu-etal-2020-unsupervised}.
Texts with high probability given an LM are considered high-quality texts.
In practice, several methods can be used to calculate the score, such as perplexity, negative log-likelihood, and SLOR~\citep{kann-etal-2018-sentence}.
For example, the Scribendi Score~\citep{islam-magnani-2021-end}, a reference-free evaluation metric for GEC based on GPT-2's perplexity scores, has been proposed.
This metric is reported to correlate better with human ratings than traditional reference-based metrics such as M2~\citep{dahlmeier-ng-2012-better} and GLEU~\citep{napoles-etal-2015-ground,napoles2016gleu}.
This approach is often combined with others, such as those described in Section~\ref{subsec:similarity}.

\subsection{Approaches focused on individual textual quality aspects}\label{subsec:heuristic}
This subsection discusses several approaches that address specific aspects of textual quality.

\subsubsection{Coherence}

Coherence evaluation methods are often trained on synthetic tasks.
For example, a synthetic task is a task that creates incoherent text (negatives) by inserting extra sentences or shuffling the order of sentences, and then classifies negatives and positives~\citep{moon-etal-2019-unified,shen-etal-2021-evaluating}.

These tasks are similar to Next Sentence Prediction and Sentence Order Prediction used in the pre-training phase of masked language models such as BERT~\citep{devlin-etal-2019-bert} and ALBERT~\citep{Lan2020ALBERT:}.
Therefore, masked language models are also used to evaluate text coherence~\citep{zhu-bhat-2020-gruen}.

However, it has been reported that even if the performance is high on synthetic tasks, performance on downstream NLG tasks can be low~\citep{mohiuddin-etal-2021-rethinking}.
\citet{steen-markert-2022-find} also investigate the performance of reference-free coherence evaluation metrics in summarization tasks and found that the correlation between human judgment and metrics scores was low.

\subsubsection{(Lack of) redundancy}
Lack of redundancy is an important quality criterion for tasks generating paragraph- and document-level text, such as summarization and story generation. 
To measure redundancy, heuristic methods such as n-gram repetition counts and superficial expression repetition counts are commonly used~\citep{zhu-bhat-2020-gruen,xiao-carenini-2020-systematically}.
However, these methods fail to capture contextual redundancies, where different surface forms have semantically similar content.

\subsubsection{Readability}
Readability assessment is the task of estimating how easy or difficult a text is to read, which has long been addressed, not only in the language processing but also in language education. 
Fluency, redundancy, and coherence, among other factors, contribute to readability.
Another important aspect of readability relates to its readers, because it is thought that expressions that are unnecessarily difficult for readers should be avoided.

Readability measures, including and Flesch Reading Ease and Flesch-Kincaid Grade Level~\citep{flesch1948new,kincaid1975derivation}, have long been used.
These measures are linear combinations of the number of words per sentence, the number of syllables per word, and so on, using carefully adjusted weights.
Flesch-Kincaid Grade Level is often used in evaluation of simplification tasks~\citep{alva-manchego-etal-2019-easse}.
However, it is reported that these methods can yield inaccurate or misleading results~\citep{tanprasert-kauchak-2021-flesch}, so caution is advised in their utilization.

\subsubsection{Diversity}
Diversity is often considered to be desirable in tasks such as dialogue, story generation, and paraphrasing. 
Most diversity evaluation metrics are surface based, based on n-grams; often used ones are Distinct-N~\citep{li-etal-2016-diversity}, Self-BLEU~\citep{sun-zhou-2012-joint,Serban_Sordoni_Lowe_Charlin_Pineau_Courville_Bengio_2017}\footnote{There are two methods called Self-BLEU; one computes the BLEU score between inputs and outputs~\citep{sun-zhou-2012-joint} and the other computes the BLEU score between generated sentence sets~\citep{Serban_Sordoni_Lowe_Charlin_Pineau_Courville_Bengio_2017}.}, and Pairwise-BLEU~\citep{pmlr-v97-shen19c-pairwise-bleu}. 
In practice, increasing diversity may sacrifice the clarity of the generated text, so it should always be used in conjunction with other evaluation criteria~\citep{ippolito-etal-2019-comparison,kulikov-etal-2019-importance}.

\section{Other uses of reference-free metrics}
\label{sec:other-usages}
We discuss other uses of reference-free metrics that go beyond their use in evaluating models.

\subsection{Rewards for reinforcement learning}
\label{subsec:reward4rl}
The reference-free approach is often used as a reward function for reinforcement learning~\citep{scialom-etal-2019-answers,cho-etal-2022-fine}. 
A reference-based metric is also sometimes used as a reward function, but it cannot be used online and is typically used to optimize a model with the objective function (evaluation metrics) for each task~\citep{bahdanau2017an}.

Reinforcement learning from human feedback (RLHF), which optimizes NLG models based on human feedback on their output to reflect human preference, has gained much attention~\citep{NIPS2017_d5e2c0ad,NEURIPS2020_1f89885d}.
In the RLHF framework, a model that outputs a scalar reward to quantify human preferences is used as the reward function, not human feedback directly.
This reward model is basically the same as that of the reference-free metrics introduced in Section~\ref{subsec:human-judgement}. 
Thus, improvement to reference-free evaluation metrics has become increasingly important for both evaluating and learning NLG models.

\subsection{Data selection for training NLG models}
Huge, high-quality data can help train better NLG models~\citep{https://doi.org/10.48550/arxiv.2001.08361}.
In recent years, approaches to crawling data on the Web~\citep{banon-etal-2020-paracrawl} and various data augmentation techniques~\citep{feng-etal-2021-survey} have been developed to increase the training data for NLG models.
These techniques can increase the training data, but these augmented data generally contain noise.
Therefore, it is important to use cleaning technology.
The data-cleaning task is essentially the same as the reference-free evaluation metric.
For example, dual conditional cross-entropy filtering~\citep{junczys-dowmunt-2018-dual} is essentially the same idea as the method introduced in Section~\ref{subsubsec:peer-review}.

In active learning, a reference-free evaluation metric is also used to select data to be requested for humans to annotate~\citep{chimoto-bassett-2022-comet,zeng-etal-2019-empirical}.
By requesting humans to annotate the output of an NLG model with low scores on the reference-free metrics can produce data to build better NLG models that are preferentially selected and labeled.

Studies of data selection can help identify some weaknesses of reference-free metrics, as they apply these metrics to different data distributions than those typically used in the evaluation of NLG evaluation metrics.
In fact, \citet{bane-etal-2022-comparison} compared some data filtering methods to find that COMET-QE is strong in detecting word order but not in detecting mismatches of numbers between context and target text.
Complementary research between data selection and reference-free metrics is becoming increasingly important.

\subsection{Reranking and filtering of model outputs}\label{subsec:reranker}
Reranking or filtering relates to the task of selecting good outputs or removing bad outputs from a set of hypothesis.
Reference-free evaluation metrics can be used for these tasks~\citep{chollampatt-ng-2018-neural}.
In comparison to standard model evaluation, reranking and filtering require often runtime evaluation, which need to be done quickly.
Therefore, efficient reference-free evaluation metrics~\citep{https://doi.org/10.48550/arxiv.2209.09593, 10.1007/978-981-19-7596-7_12} are valuable for these functions.

\section{Discussion}
\label{sec:challenges-future-research}

\subsection{Reference-free vs. reference-based}
In this section, we summarize the main characteristics of reference-free and reference-based metrics.

Faced with the choice between reference-based and reference-free methods, an important consideration is related to the \emph{cost} of the method.
Reference-based approaches require references, which are often expensive to collect.
In particular, it is not easy to collect enough references for open-ended tasks, such as dialogue and story generation. 
Reference-free metrics do not require references.
However, as pointed out in Section~\ref{sec:generic-approach}, many reference-free methods require training data, such as human judgments, or parallel data. 
Consequently, a reference-free approach is by no means cost-free.

The second consideration is \emph{performance}. 
Some researchers have reported that reference-free metrics produce lower evaluation performance than reference-based ones~\citep{banchs-li-2011-fm,fonseca-etal-2019-findings}, but in recent years, some studies have reported the reverse.
For example, \citet{kasai-etal-2022-bidimensional} found that reference-free metrics perform better than reference-based ones when there are few references. 
If the reference quality is poor, the evaluation performance of the reference-based method suffer (see Table~13 of \citet{freitag-etal-2021-results}).
Reference-free metrics are thus not always inferior to reference-based metrics.

\subsection{LLMs and reference-free evaluation metrics}

LLMs have altered the NLG evaluation landscape in broadly two ways.

On the one hand, LLMs have improved the performance of NLG models.
As a result, texts generated by these NLG models are often indistinguishable from those written by humans~\cite{clark-etal-2021-thats}. 
These improvements present a challenge for the evaluation of such texts because, in some cases, the errors and infelicities that separate competing models are becoming increasingly subtle.

On the other hand, with the advent of LLMs, reference-free metrics have also advanced. 
For example, in addition to the methods presented in Section~\ref{subsubsec:llm-as-a-judge}, LLMs are used as regression models in Section~\ref{subsec:human-judgement}~\citep{guerreiro2023xcomet}. 
It would also be possible to use LLMs as NLG models in the methods presented in Section~\ref{subsubsec:sim} and~\ref{subsubsec:peer-review} in the creation of pseudo-ratings presented in Section~\ref{subsubsec:psuedo-rating}.\footnote{See the survey by \citet{li2024leveraging} for evaluation metrics using LLMs.}

\subsection{Future research: Combining different approaches to evaluation}

In future, we expect that different approaches to evaluation, which have complementary strengths, will be combined to achieve superior results. 
For example, it has been suggested that the types of errors that humans can easily detect differ from those that automatic evaluation metrics can easily identify~\citep{chen-etal-2024-humans}.
By combining human and automatic evaluation, it may therefore be possible to reduce the workload of human evaluators~\citep{zhang-etal-2021-human}, or to achieve results with superior validity and reliability. 
For example, human evaluation could focus on cases in which discrepancies occur across evaluation metrics.
We believe that exploring effective and efficient ways to combine human evaluation and automatic evaluation metrics is an important direction for future research.

\section{Conclusion}
Evaluation is an essential step for the development of NLG models of any kind\citep{albert-etal-2018-survey-nlg, van-der-lee-etal-2019-best, Reiter-book}.
In this survey, we have discussed existing computational, \emph{reference-free} evaluation metrics for a broad range of NLG tasks, including not only data-to-text NLG but also such text-to-text tasks as machine translation and question answering, for example, which are not always included in discussions of NLG and NLG evaluation. 
These reference-free metrics have recently attracted more attention and shown improved performance. 

We have shown that although a wide range of reference-free metrics have been proposed, all of these can be seen as variations on a few basic themes, such as learning from human judgments; exploiting the correspondence between the context and the hypothesis (as these notions are defined in section~\ref{sec:terminology}); and using peer evaluation. 
We have also outlined the assumptions and limitations underlying each method.

We hope that this survey will lead to further research on reference-free metrics, and that this will lead to further insights in the question of when these metrics are most useful, and how they are best combined with other approaches to evaluation.

\clearpage

\bibliography{custom}

\appendix

\end{document}